\documentclass[11pt]{article}

\usepackage[preprint]{acl}

\usepackage{times}
\usepackage{latexsym}
\usepackage[T1]{fontenc}
\usepackage[utf8]{inputenc}

\usepackage{inconsolata}
\usepackage{graphicx}
\usepackage{subfigure}
\usepackage{booktabs} 
\usepackage{paralist}

\usepackage[table]{xcolor}
\usepackage[most]{tcolorbox}
\usepackage{tabularx}
\usepackage{colortbl}

\usepackage{algorithm}
\usepackage{algorithmic}
\usepackage{amsmath}
\usepackage{amssymb}
\usepackage[capitalise]{cleveref}
\usepackage{hyperref}
\usepackage{bbm}
\usepackage{multirow}
\usepackage[most]{tcolorbox}
\usepackage{quoting}
\usepackage{mdframed}
\usepackage{enumitem}
\usepackage{microtype}

\crefname{section}{Section}{Sections}
\crefname{subsection}{Section}{Sections}
\crefname{subsubsection}{Section}{Sections}
\crefname{figure}{Figure}{Figures}
\crefname{table}{Table}{Tables}
\crefname{equation}{Eq.}{Eqs.}
\crefname{appendix}{Appendix}{Appendices}
\crefname{algorithm}{Algorithm}{Algorithms}
\crefname{enumi}{Item}{Items}

\newtcolorbox{UserPrompt}{
  enhanced,
  colback=gray!10,
  colframe=blue!80!black,
  coltitle=white,
  fonttitle=\bfseries,
  title=User Prompt,
  arc=3pt,
  boxrule=1pt,
  left=2mm,
  right=2mm,
  top=1mm,
  bottom=1mm
}

\newtcolorbox{SystemPrompt}{
  enhanced,
  colback=green!5!white,
  colframe=green!75!black,
  coltitle=white,
  fonttitle=\bfseries,
  title=System Prompt,
  arc=3pt,
  boxrule=1pt,
  left=2mm,
  right=2mm,
  top=1mm,
  bottom=1mm
}

\newtcolorbox{FullWidthBox}[1]{
  enhanced,
  colback=blue!2!white,
  colframe=blue!75!black,
  coltitle=white,
  fonttitle=\bfseries\large,
  title=#1,
  arc=4pt,
  boxrule=1pt,
  width=\textwidth,
  left=4mm,
  right=4mm,
  top=3mm,
  bottom=3mm,
  before skip=15pt,
  after skip=15pt
}

\mdfdefinestyle{quotestyle}{
    linewidth=2pt,
    linecolor=gray,
    topline=false,
    bottomline=false,
    rightline=false,
    leftmargin=0pt,
    innerleftmargin=10pt
}

\newcommand{\oss}{GPT-OSS-120B}
\newcommand{\qwenthreetwo}{Qwen3-32B}
\newcommand{\glmthreetwo}{GLM-4-32B-0414}
\newcommand{\llamaseventy}{Llama-3.3-70B-Instruct}
\newcommand{\gemmatwoseven}{Gemma-3-27B}
\newcommand{\exaonethreetwo}{Exaone-4-32B}

\newcommand{\gemmanine}{Gemma-2-9B-It}
\newcommand{\qweneight}{Qwen3-8B}
\newcommand{\exaoneeight}{Exaone-3.5-7.8B-Instruct}
\newcommand{\deepseekmath}{Deepseek-Math-7B-Instruct}
\newcommand{\glmnine}{GLM-4-9B-Chat}
\newcommand{\yinine}{Yi-1.5-9B-Chat-16K}

\newcommand{\geminitwofiveflash}{Gemini-2.5-Flash} 
\newcommand{\geminithreeflash}{Gemini-3-Flash} 
\newcommand{\qwenembed}{Qwen3-Emedding-8B}

\newcommand{\method}{\textsc{CASCAL}}
\newcommand{\methodlong}{\underline{C}onsensus-\underline{A}ware \underline{S}kill \underline{C}lustering and \underline{A}ggregation for \underline{L}LMs
}
\newcommand{\methodgt}{\textsc{CASCAL-GT}}
\newcommand{\poolsmall}{\textsc{Pool-Small}}
\newcommand{\poollarge}{\textsc{Pool-Large}}
\newcommand{\settinglong}[0]{Routing with Generated Data}
\newcommand{\setting}[0]{\textsc{RGD}}

\newcommand{\smoothie}{\textsc{Smoothie-Train}}
\newcommand{\avengers}{\textsc{Avengers}}
\newcommand{\llmrank}{\textsc{LLMRank}}

\newcommand{\myparagraph}[1]{\vspace{0.25em}\noindent \textbf{#1}\hspace{0.5em}}

\definecolor{forestgreen}{RGB}{34, 139, 34}

\title{Routing with Generated Data: Annotation-Free \\ LLM Skill Estimation and Expert Selection}

\author{Tianyi Niu$^1$, Justin Chih-Yao Chen$^1$, Genta Indra Winata$^2$, Shi-Xiong Zhang$^2$, \\ 
\textbf{Supriyo Chakraborty$^2$}, \textbf{Sambit Sahu$^2$}, \textbf{Yue Zhang$^1$}, \\
\textbf{Elias Stengel-Eskin$^3$}, \textbf{Mohit Bansal$^1$}\\
$^1$UNC Chapel Hill$\quad$$^2$Capital One$\quad$$^3$The University of Texas at Austin}

\begin{document}
\maketitle

\begin{abstract}
Large Language Model (LLM) routers dynamically select optimal models for given inputs. Existing approaches typically assume access to ground-truth labeled data, which is often unavailable in practice, especially when user request distributions are heterogeneous and unknown. We introduce Routing with Generated Data (RGD), a challenging setting in which routers are trained exclusively on generated queries and answers produced from high-level task descriptions by generator LLMs. We evaluate query-answer routers (using both queries and labels) and query-only routers across four diverse benchmarks and 12 models, finding that query-answer routers degrade faster than query-only routers as generator quality decreases. Our analysis reveals two crucial characteristics of effective generators: they must accurately respond to their own questions, and their questions must produce sufficient performance differentiation among the model pool. We then show how filtering for these characteristics can improve the quality of generated data. We further propose \method{}, a novel query-only router that estimates model correctness through consensus voting and identifies model-specific skill niches via hierarchical clustering. \method{} is substantially more robust to generator quality, outperforming the best query-answer router by 4.6\% absolute accuracy when trained on weak generator data.\footnote{Code is available at~\url{https://github.com/tianyiniu/RoutingGenData}.}
\end{abstract}

\section{Introduction}
\label{intro}

\begin{figure}[t]
    \centering
    \includegraphics[width=\columnwidth]{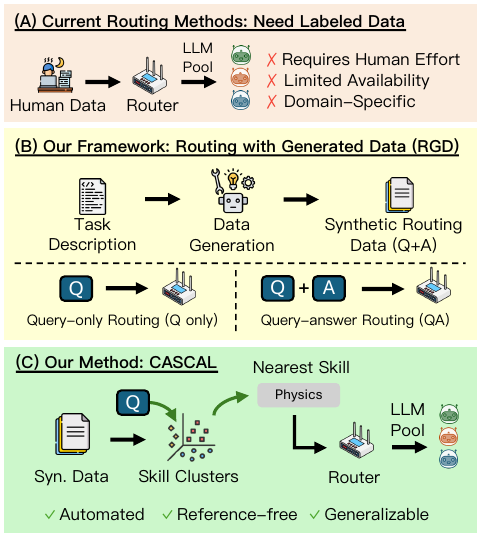}
    \caption{Overview of Routing with Generated Data (RGD). (A) Most existing routing methods require human-labeled data for skill estimation and expert selection. 
    (B) RGD generates routing data from task descriptions, enabling both query-only and query-answer routing.
    (C) \method{} extracts skills clusters from generated queries and routes to models without ground-truth labels.}
    \label{fig:teaser}
\end{figure}

\begin{figure*}[t]
    \centering
    \includegraphics[width=2\columnwidth]{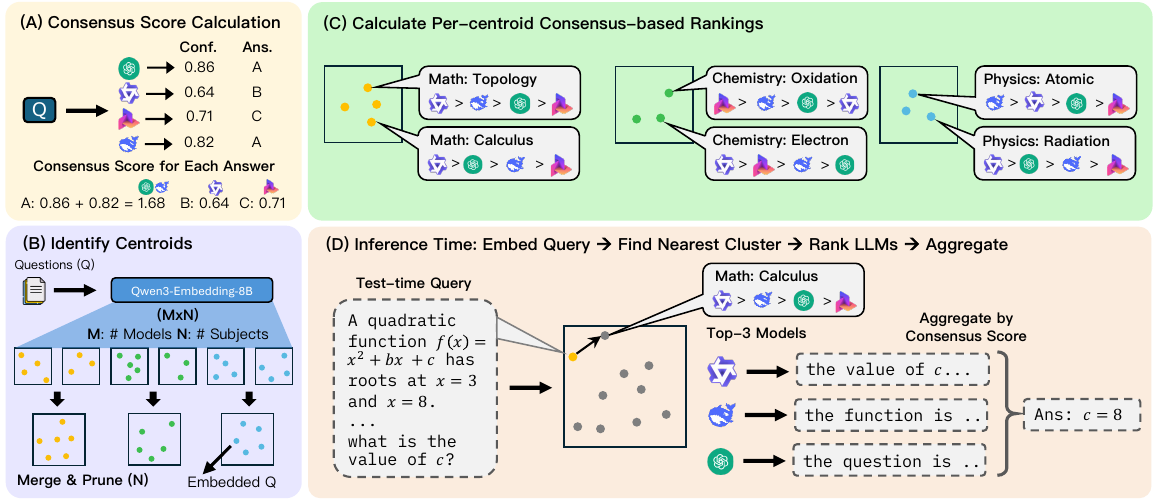}
    \caption{Overview of \method{}. (A) Consensus Scoring: we extract model responses for each query and compute confidence-weighted consensus scores. (B) Centroid Identification: For each model and subject, we cluster queries where the model demonstrates proficiency to obtain skill centroids, then we merge similar centroids across models. (C) Cluster Ranking: we assign queries to their nearest centroid and rank models within each cluster by average consensus score. (D) Inference: we route test queries to the nearest subject and centroid, select the top-3 (or top-1) ranked models, and aggregate responses via consensus voting.}
    \label{fig:cascal}
\end{figure*}

As the ecosystem of large language models (LLMs) matures, there are an increasing number of open-weight LLMs available to the public \cite{opensoucesurvey}. While leaderboards provide aggregate pictures of model performance, a growing body of work in LLM routing has proposed ways to \emph{dynamically} determine which model(s) to select for a given \emph{instance}. At its core, routing relies on recognizing each model's strengths and weaknesses at a fine granularity and routing queries accordingly, i.e., estimating each model's skills and selecting expert models.
Most prior routing methods generally follow one of three paradigms: (1) training a classifier to select models given an input \cite{agrawal2025llmrank, Ong2025routellm, tsiourvas2025causalllmrouting, chen2024frugalgpt}, (2) profiling models using inferred natural language skills \cite{chen2025symbolic, chen2025tagrouter, maimon2025iqtestllmsevaluation, shah2025selectthenroute, dong2025sdag}, or (3) clustering-based approaches that identify skills in the embedding space \cite{zhang2025advengers, jitkrittum2025googleroute, pichlmeier2024expertrouter}. Crucially, these approaches all \emph{assume access to labeled, in-domain data}, either for training and validating the classifier, or for inferring and validating skills. However, in real-world settings where the distribution of user requests is heavily heterogeneous and not known \textit{a priori}, this assumption may not hold, and data may be difficult to obtain. In other words, routing decisions may need to be made \emph{without access to any ground-truth in-domain data} (\cref{fig:teaser}A). 

To capture this scenario, we lift the assumption of ground-truth data and introduce \textbf{\settinglong{} (\setting{})}, a challenging new routing setting (\cref{fig:teaser}). \setting{} requires task-driven data generation, for which we provide a series of data domains, each with a natural language domain description, a pool of candidate models, and a held-out test set. Rather than providing in-domain data (as traditional routing settings might have), \setting{} generates query-answer data pairs that seek to differentiate between strong and weak models. The generated queries -- and, depending on the routing method, their answers -- are then used to fit different routers to the model pool (see \cref{fig:teaser}B).

Given the demonstrated strength of LLMs in generating data \cite{numinamath, khan2025dataenvgym, long2024datagensurvey, Nadas2025}, we focus specifically on LLM-based data generation, addressing the following key questions: 

\begin{enumerate}[noitemsep, topsep=0pt]
    \item \emph{RQ1: How do routing methods designed for real data adapt to generated data in \setting{}?}
    \item \emph{RQ2: What are the characteristics of strong profiling data in \setting{}?}
    \item \emph{RQ3: What factors can improve the quality of generated profiling data?}
\end{enumerate}

Our analysis reveals two key insights: (1) weaker generators produce unreliable answer labels but can still generate useful queries, and (2) effective routing requires identifying fine-grained skill niches where models differ. Building on these findings, we develop a novel label-free routing method: \methodlong{} (\method{}) (see \cref{fig:teaser}C and \cref{fig:cascal}).

\method{} is characterized by two core ideas: 
\textbf{(1) Consensus-based correctness estimation:} Given that majority vote is often a strong signal for correctness \cite{wang2022self, chen-etal-2024-reconcile}, we calculate a confidence-weighted majority vote, thereby giving us a continuous metric that captures how closely each specific model is aligned with the majority opinion for a given input (\cref{fig:cascal}A).\footnote{This metric can also be understood as a form of self-consistency across multiple models.} \textbf{(2) Hierarchical clustering to identify niche skills:} To better capture model differences on niche topics, we propose a hierarchical clustering approach that groups queries into skill clusters and identifies niche skills for each model (\cref{fig:cascal}B). For each cluster centroid, we rank models by their average consensus score across all queries assigned to that cluster (\cref{fig:cascal}C). At test time, a new query is routed to its nearest centroid, and the top-3 models for that cluster are selected as experts (\cref{fig:cascal}D).

We evaluate \method{} and other routing baselines on two model pools -- one composed of larger models (>20B parameters), and one composed of smaller models (<10B) -- and across diverse benchmarks, testing on MMLU-Pro \cite{mmlupro}, MedMCQA \cite{medmcqa}, SuperGPQA \cite{supergpqa}, BigBench-ExtraHard \cite{bbeh}.
We contrast \method{} with competitive query-answer routing baselines (i.e., methods that use both queries and answers) like \llmrank{} \cite{agrawal2025llmrank} and \avengers{} \cite{zhang2025advengers}, as well as another strong query-only baseline, \smoothie{} \cite{guha2024smoothie}. 

First, addressing \emph{RQ1}, we find that, while most routing methods perform similarly with ground-truth data, existing answer-based routing methods perform poorly in the \setting{} setting. For example, with data generated by \exaoneeight{}, \method{} achieves an average accuracy of 61.1\%, whereas \llmrank{} and \avengers{} achieve 57.1\% and 58.9\%, respectively with the large model pool.
Second, for \emph{RQ2}, we find that query-only methods are more robust to data quality, and that weaker LLMs are more adept at generating routing queries than generating answers to those queries. 
Finally, addressing \emph{RQ3}, we use data filtering to assess which qualities of the generated data transfer well. 
We find that filtering for variance-inducing questions with high consensus by stronger models produces more informative routing samples and improves \method{}'s performance.
Taken together, our work opens up the possibility of estimating the strengths of a diverse model pool \emph{without} access to ground-truth annotations.

\section{Routing Formulation under \setting{}}
\label{sec:setting}

We formalize the routing problem in the \setting{} setting as follows. 
Let $\mathcal{M} = \{m_1, \ldots, m_M\}$ be a pool of $M$ candidate LLMs. 
Given a query $q$ from a domain $\mathcal{D}$, the goal of a router $\pi$ is to select a subset of $k$ models from $\mathcal{M}$ that maximizes performance on $q$. 
Formally, the router learns a mapping:
\vspace{-0.5em}
\begin{equation}
\pi: q \rightarrow \{m_{i_1}, \ldots, m_{i_k}\} \subseteq \mathcal{M},
\vspace{-0.25em}
\end{equation}
where the chosen models may be used alone ($k=1$) or aggregated via ensemble methods such as majority voting or via a separate aggregator model.

\subsection{Training and Inference Protocols}
During training, routing methods construct internal representations of model capabilities using a training dataset $\mathcal{D}_{\text{train}} = \{(q_i, a_i^*, \{r_i^{(j)}\}_{j=1}^M)\}_{i=1}^N$, where $q_i$ is a query, $a_i^*$ is an answer, and $r_i^{(j)}$ is the response from model $m_j$. Methods differ in what information they utilize:
\begin{itemize}[noitemsep,topsep=0pt,leftmargin=*]
    \item \textbf{Query-answer methods} assume access to the full tuple $(q, a^*, \{r^{(j)}\})$, using ground-truth labels to compute correctness signals.
    \item \textbf{Query-only methods} assume access only to $(q, \{r^{(j)}\})$, forgoing ground-truth labels $a^*$.
\end{itemize}

At inference time, given a test query $q_{\text{new}}$, the router selects the best model(s) based on the capabilities inferred during training. 

\subsection{\settinglong{} (\setting{})}
In \setting{}, routers are trained on model-generated data. We construct datasets by prompting a generator model to produce a training dataset $\{\hat{q}_i, \hat{a}_i\}$ from a set of task descriptions that briefly highlight each task's primary topics. Given a description $d_t$, the generator model $M_{\text{gen}}$ generates a single output containing a query-answer pair conditioned on these descriptions: $(q, a)_{\text{synthetic}} = M_{\text{gen}}(d_t)$. From this, we can obtain $\hat{\mathcal{D}}_{train}$ by adding model responses from the model pool, with which we can train a router. 

\section{Routing Methodology: \method{}}
\label{sec:method}

To tackle the challenges of \setting{}, we propose \method{} (shown in \cref{fig:cascal}), a consensus-based query-only router. The method builds on two core insights: (1) majority consensus among models serves as a reliable proxy for correctness, and (2) hierarchical clustering can identify model-specific ``niche'' skills where individual models excel. Unlike query-answer routing methods that require labeled data to compute correctness signals, \method{} derives quality estimates entirely from model agreement patterns, making it well-suited to RGD where generated answers may be unreliable.

\myparagraph{Consensus Score Calculation.}
Given a query $q$ and responses from a pool of models $\mathcal{M} = \{m_1, \dots, m_M\}$, we estimate model accuracy via confidence-weighted consensus (see \cref{fig:cascal}A). For each model $m_j$, we normalize its log-probability scores across the dataset to obtain a standardized confidence score $Z_{i,j}$ for query $q_i$. The consensus score for model $m_j$ on query $q_i$ is:
\begin{equation}
    C_{i,j} = \sum_{k=1}^{M} \mathbb{I}(a_{i,j} = a_{i,k}) \cdot Z_{i,k}
\end{equation}
where $a_{i,j}$ is the answer extracted from model $m_j$'s response $r_{i,j}$. This rewards answers supported by multiple high-confidence models.

\myparagraph{Skill-centric Routing via Clustering.}
\method{} captures model specialization by constructing skill clusters within each task (see \cref{fig:cascal}B). During training, for each task $t$ and corresponding set of queries $Q_t$, we first identify queries for each model $m$ aligned with the consensus $a^{maj}_i$:

\begin{align}
    Q^{\text{strong}}_{m,t} &= \left\{ q_i \in Q_t : a_{i,m} = a^{maj}_i \right\}. \\
    a^{maj}_i &= \arg\max_{a} \sum_{k=1}^{M} \mathbb{I}(a_{i,k} = a).
\end{align}
$Q^{\text{strong}}_{m,t}$ thus represents the set of queries in which model $m$ demonstrates proficiency. Using k-means on this set of queries, we then cluster embeddings of $Q^{\text{strong}}_{m,t}$ to obtain skill centroids for task $t$ (e.g., ``calculus'' vs. ``topology''). Centroids that are too close or yield identical model rankings are merged or pruned (see \cref{fig:cascal}C). Finally, we assign each training query to its nearest centroid---thereby forming skill clusters---and rank models within each cluster by their average consensus scores. This yields a mapping from skill clusters to ranked expert models. Implementation details are provided in \cref{sec:cascal}.

\myparagraph{Inference-time Routing and Aggregation.}
At inference time, \method{} routes an input query through three stages: (1) the query is assigned to the most relevant task based on embedding similarity; (2) within that task, the query is matched to its nearest skill cluster; and (3) the top-ranked models associated with that cluster are selected and their outputs are aggregated via consensus scoring (see \cref{fig:cascal}D). The final answer is chosen as the one with the highest aggregated score.\footnote{Note that aggregating via consensus score is very similar to majority voting in practice. The only case where the two aggregation methods will diverge is when all three selected models provide different answers. In that case, the most confident model's answer will be selected.}

\myparagraph{\method{} Variants.}
In addition, we evaluate two variants of \method{}. First, \textbf{\method{} (Top-1)} omits the aggregation stage and directly routes to the single top-ranked model ($K=1$), enabling comparison against routers that select a single model per query. Second, \textbf{\methodgt{}} replaces consensus scores with a ground-truth label for ranking models within clusters during training, making \methodgt{} a query-answer router. Details are provided in \cref{sec:cascal}.

\begin{table}[t]
\centering
\resizebox{\linewidth}{!}{
\begin{tabular}{llc}
\toprule
\textbf{Pool} & \textbf{Model} & \textbf{Params} \\
\midrule
\multirow{6}{*}{\shortstack{POOL-\\LARGE}} 
 & GPT-OSS~\cite{agarwal2025gpt}          & 120B \\
 & LLaMA-3.3~\cite{llama3_3_70b}    & 70B \\
 & Qwen-3~\cite{yang2025qwen3technicalreport}        & 32B \\
 & GLM-4~\cite{glm4-32b-0414}        & 32B \\
 & Exaone-4~\cite{bae2026exaone4}     & 32B \\
 & Gemma-3~\cite{gemmateam2025gemma3}      & 27B \\
\midrule
\multirow{6}{*}{\shortstack{POOL-\\SMALL}}
 & Gemma-2~\cite{gemmateam2024gemma2}       & 9B \\
 & GLM-4~\cite{glm2024}         & 9B \\
 & Yi-1.5~\cite{ai2025yiopenfoundationmodels}            & 9B \\
 & Qwen-3~\cite{{yang2025qwen3technicalreport}}         & 8B \\
 & Exaone-3.5~\cite{an2026exaone35serieslarge}        & 7.8B \\
 & DeepSeek-Math~\cite{shao2024deepseekmath}    & 7B \\
\bottomrule
\end{tabular}
}
\caption{Model pools used in routing experiments. We evaluate routing methods on
two disjoint model pools grouped by parameter scale.}
\label{tab:model_pools}
\end{table}

\section{Experimental Results}

\subsection{Experiment Setup}
\myparagraph{Model Pools.}
\label{sec:models_datasets_description}
We evaluate routing methods on two disjoint model pools, summarized in
Table~\ref{tab:model_pools}.
\poollarge{} contains six large-scale models with more than 20B parameters,
while \poolsmall{} consists of six smaller models with fewer than 10B parameters.
This design allows us to study routing behavior under different model capacity
regimes. We use \qwenembed{} \cite{zhang2025qwen3embed} to obtain embeddings for clustering.

\myparagraph{Datasets and Tasks.}To comprehensively examine how generated data affects routing quality, we construct scenarios using \setting{} across four distinct datasets: \textbf{MMLU-Pro} \citep{mmlupro}, \textbf{MedMCQA} \citep{medmcqa}, \textbf{SuperGPQA} \citep{supergpqa}, and \textbf{BigBench-Extra-Hard} \citep[BBEH;][]{bbeh}. We subdivide each dataset into tasks; for MMLU-Pro, MedMCQA, SuperGPQA, each task represents a subject. 
We note that -- as with common voting methods like Self-Consistency \citep{wang2022self} -- \method{} requires tasks to have a discrete set of output classes for consensus and aggregation, so we remove tasks from BBEH that are difficult to fit into a discrete format (see \cref{app:dataset_details}). We divide each dataset into 6:4 train-test splits, stratified by tasks.  

\myparagraph{\setting{} Data Generation.} 
\label{sec:rgd_description}
We evaluate routing methods across four \setting{} scenarios, each defined by a different generator model $M_{\text{gen}}$: Real data, \geminitwofiveflash{}, \qwenthreetwo{}, and \exaoneeight{}.
These models were chosen to represent one frontier LLM, one stronger open-weight model, and one smaller 8B parameter model. 
First, for each scenario, we obtain a set of task descriptions by prompting a descriptor model to summarize a given task $t$ using a small number of seed examples\footnote{We use \geminitwofiveflash{} as the descriptor model with 5 in-context queries for all experiments. This is for efficiency; task descriptions can also be human-written.}. We include all task descriptions in Appendix~\ref{sec:task_descriptions} and prompts in Appendix~\ref{sec:model_response_prompt}. We manually verify that the generated descriptions do not leak validation examples. Next, we independently prompt each generator to generate 5,000 queries $\hat{q}$ and answers $\hat{a}$ per domain.

\subsection{Baselines} 
\label{sec:baseline_description}
\myparagraph{Learning-free Baselines.} 
\textbf{Top-1} selects the single best-performing model as measured by average validation set performance. 
\textbf{Top-3 Vote} applies majority voting over the top three models. 
Note that these set a ceiling on performance in the \setting{} setting, as they use real validation data to determine the top-1 and top-3 models.
\textbf{Random-1} samples a model from the pool, and \textbf{Random-3 Vote} applies majority voting over three sampled models. 

\myparagraph{Query-answer Routers.}
\label{sec:qa_routers_description}
These routing methods that assume access to ground-truth labels during training represent the standard paradigm in LLM routing, where correctness signals are used to learn model strengths. We evaluate the following representative approaches:

\begin{itemize}[noitemsep,topsep=0pt,leftmargin=*]
    \item \textbf{\llmrank{}} \citep{agrawal2025llmrank}: Extracts interpretable features from prompts (task type, complexity, domain, etc.) and trains a neural ranking model to predict per-model capability scores. We document our features used in \cref{app:dataset_details} and denote our implementation as \llmrank{}.
    \item \textbf{\avengers{}} \citep{zhang2025advengers}: Embeds queries, clusters them via $k$-means ($k=64$), and scores each model's accuracy per cluster. At test time, queries are routed to the nearest cluster, the top-3 models are selected, and their outputs are aggregated via a majority vote.
\end{itemize}

For all query-answer routers, training involves computing correctness signals $\mathbbm{1}[r_i^{(j)} = a^*_i]$ for each model response and using these to fit routers. 
In the \setting{} setting, this becomes $\mathbbm{1}[r_i^{(j)} = \hat{a}_i]$, where $\hat{a}_i$ is a model-generated answer.

\myparagraph{Query-only Routers.}
\label{sec:q_only_router_description}
Query-only routers operate without ground-truth labels and are particularly suited to RGD settings as generated data may lack reliable answer labels. We also consider \textbf{\smoothie{}} \citep{guha2024smoothie}, a weak supervision-inspired approach which models embedding differences between LLM outputs as a multivariate Gaussian to derive per-model quality scores. Since we evaluate routing under domain shift, we use \smoothie{} variant of the algorithm. As \method{} by default aggregates 3 models, we further show results from majority voting using the top-3 ranked by \smoothie{} (Top-3). For query-only methods, training involves computing proxy quality metrics from model responses $\{r_i^{(j)}\}_{j=1}^M$ without reference to ground-truth.

\subsection{Main Results and Discussions}

\subsubsection{RQ1: How do routing methods adapt to the \setting{} setting?}
\label{sec:rq1}

\begin{figure*}[t]
    \centering
    \includegraphics[width=\textwidth]{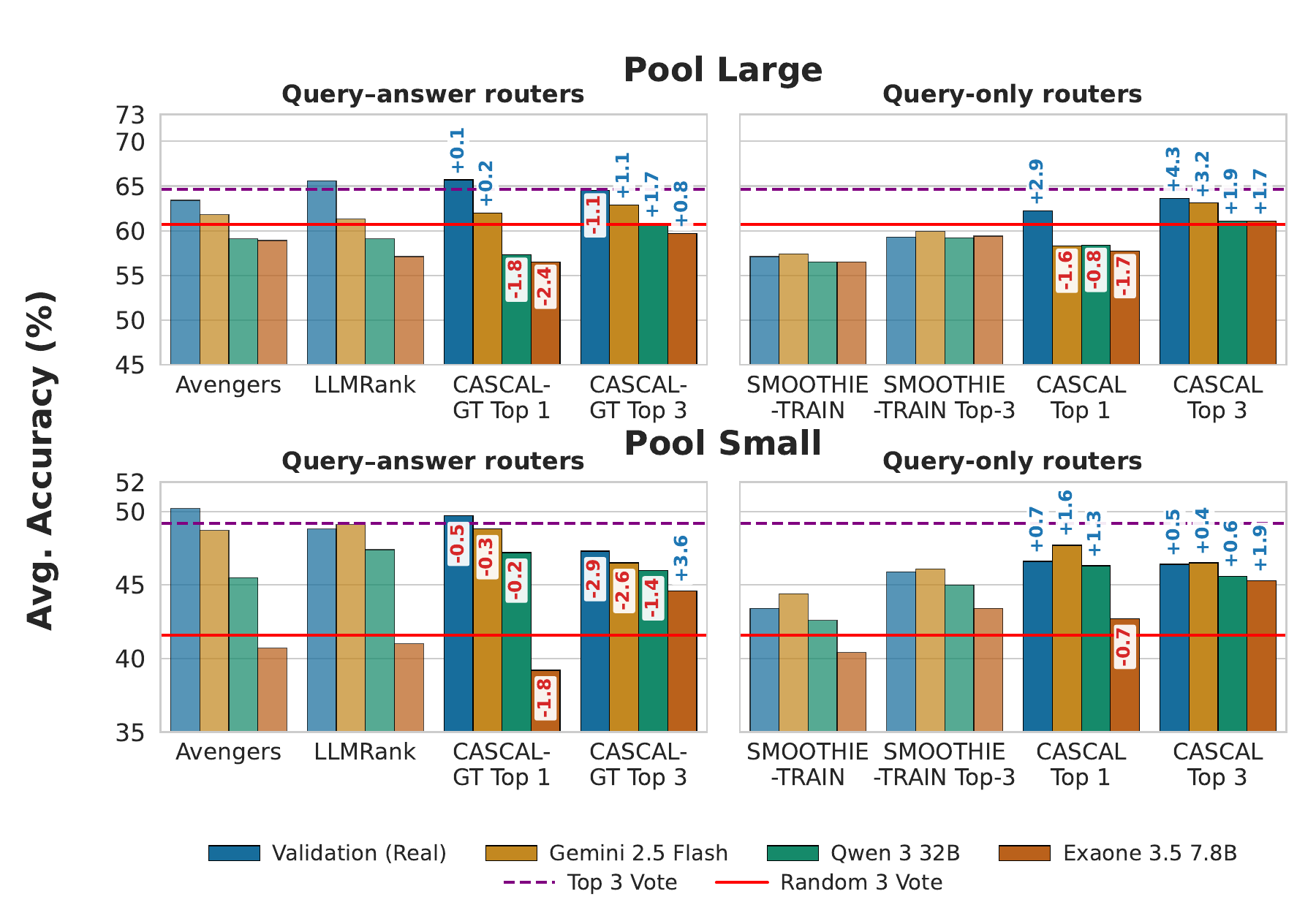}
    \caption{Routing accuracy across RGD scenarios for Pool-Large (left) and Pool-Small (right). Colors indicate the source of routing data: validation data or data generated by different LLMs. Each bar represents the router’s average test accuracy across four datasets (MMLU-Pro, SuperGPQA, MedMCQA, and BigBench-Extra-Hard). Annotations indicate the absolute accuracy improvement of CASCAL variants over the strongest non-CASCAL baseline within the same routing family (subplot) under the same data source (color).}
    \label{fig:main_bar}
\end{figure*}

We present our main results in \cref{fig:main_bar}, 
which compares the average performance of different routing methods across all \setting{} scenarios for both model pools, with three key observations. We present detailed results for all routers in \cref{sec:further_results}.

\paragraph{Stronger generators consistently produce better routers.} Across nearly all routing methods, stronger generators produce training data that yields superior routing performance. Weak generators never outperform stronger ones and generated data almost never outperforms ground-truth data, with some rare exceptions for query-only methods trained on Gemini-generated data. In \poolsmall{}, \method{} Top-1 improves from 46.6\% to 47.7\% (+1.1\%), \smoothie{} Top-1 improves from 43.4\% to 44.4\% (+1.0\%).\footnote{All percentage differences are absolute.} 
This suggests that high-quality generated data can occasionally match or exceed real data for fitting routers, particularly for query-only methods that do not rely on potentially noisy generated labels.  

\paragraph{\method{} is more robust to generator degradations.} Among query-only methods, \method{} consistently outperforms \smoothie{} across all RGD scenarios in both model pools. On \poollarge{}, \method{} achieves 63.1\%, 61.1\%, and 61.1\% accuracy under Gemini, Qwen, and Exaone generators respectively, compared to \smoothie{} Top-3's 59.9\%, 59.2\%, and 59.4\%. On \poolsmall{}, \method{} similarly beats \smoothie{} Top-3 across all scenarios (46.5\% vs 46.1\% for Gemini, 45.6\% vs 45.0\% for Qwen, and 45.3\% vs 43.4\% for Exaone). Critically, on \poollarge{}, \method{} is the only method to consistently match or exceed the Random-3 Vote baseline across all RGD scenarios. With Exaone-generated data on \poolsmall{}, query-answer methods like \avengers{} (40.7\%) and \llmrank{} (41.0\%) fall to or below the Random-3 Vote baseline of 41.6\%, while \method{} (45.3\%) maintains a substantial margin above it. This makes \method{} valuable in budget-constrained cases where only weak generators are available.

\paragraph{Weak generators disproportionately harm query-answer routers.} We observe a clear asymmetry between query-answer and query-only methods when generator quality degrades. On \poollarge{}, query-answer methods suffer substantial accuracy losses between validation and Exaone-generated data: \llmrank{} drops 8.5\% (65.6\% $\rightarrow$ 57.1\%), \methodgt{} Top-1 drops 9.2\% (65.7\% $\rightarrow$ 56.5\%), and \avengers{} drops 4.5\% (63.4\% $\rightarrow$ 58.9\%). In contrast, query-only methods exhibit much higher stability: \method{} drops only 2.5\% (63.6\% $\rightarrow$ 61.1\%), while \smoothie{} Top-3 actually remains consistent (59.3\% $\rightarrow$ 59.4\%). This pattern is even more pronounced on \poolsmall{}, where query-answer methods experience severe degradation: \avengers{} drops 9.5\% (50.2\% $\rightarrow$ 40.7\%), \methodgt{} Top-1 drops 10.5\% (49.7\% $\rightarrow$ 39.2\%), and \llmrank{} drops 7.8\% (48.8\% $\rightarrow$ 41.0\%). Similarly, query-only methods remain far more resilient: \method{} drops only 1.1\% (46.4\% $\rightarrow$ 45.3\%), maintaining its performance even under the weakest generator. These results demonstrate that query-only methods are more robust to generated data quality, making them better suited to \setting{} settings with unreliable labels from weaker generators.

\subsubsection{RQ2: What are the characteristics of strong profiling data in \setting{}?}
\label{sec:rq2}

\begin{table}[ht]
\centering
\small 
\begin{tabular}{lcc}
\toprule
\textbf{Generator} & \textbf{MMLU-Pro} & \textbf{MedMCQA} \\ 
\midrule
Exaone-3.5 7.8B & 65.6 & 75.4 \\   
\qwenthreetwo{} & 75.1 & 79.0 \\  
\bottomrule
\end{tabular}
\caption{Generator label quality measured using agreement with \geminithreeflash{} answers. The stronger generator (\qwenthreetwo{}) has consistently higher agreement.}
\label{tab:acc_wrt_gemini}
\end{table}

Here, we analyze why stronger generators lead to better routers, both from the answer and query side. We show (1) that weak generators struggle to answer their own queries, and (2) queries from stronger generators induce better model rankings. 

\myparagraph{Weaker generators struggle to answer their own queries.} For each generated query, we compare the generated answer provided by the generator model (either \exaoneeight{} or \qwenthreetwo{}) against an answer obtained using \geminithreeflash{} as a ``silver'' reference. 
Here, we make the assumption that \geminithreeflash{} is a much stronger QA model (i.e., a better approximation to a ``ground-truth'' label) compared to the generator models on MMLU-Pro and MedMCQA; this is based on its high validation accuracy on these datasets (86\% and 97\%, respectively). To further ensure the quality of Gemini's output, we generate 3 responses per query and take the majority vote response. We exclude BBEH and SuperGPQA from this analysis as \geminithreeflash{} itself cannot reliably answer the queries (\cref{app:gemini_alignment}). This experiment allows us to estimate each generator's answer quality and understand how label errors propagate to router performance.

\myparagraph{Results.} 
In \cref{tab:acc_wrt_gemini}, we find that weaker generators produce substantially less accurate labels on queries they themselves generated. \exaoneeight{} achieves only 65.6\% agreement on MMLU-Pro and 75.4\% on MedMCQA with \geminithreeflash{}. In contrast, \qwenthreetwo{} achieves higher agreement across the board: 75.1\% on MMLU-Pro and 79.0\% on MedMCQA. These errors result in noise for query-answer routers, leading to degraded performance compared to query-only routers.

\begin{table}[t]
\begin{center}
\setlength{\tabcolsep}{2.5pt}
\small
\begin{sc}
\begin{tabular}{lccccc}
\toprule
Gen. & MMLU & SGPQA & MMCQA & BBEH & Avg. \\
\midrule
\multicolumn{6}{c}{\cellcolor{gray!25}\textit{Pool Large}} \\
\midrule
Gem.  & .47 & .33 & .60 & -.20 & .30 \\
Qwen  & .47 & .47 & .07 & -.89 & .03 \\
Exa.  & -.07 & -.33 & .33 & -.47 & -.14 \\
\midrule
\multicolumn{6}{c}{\cellcolor{gray!25}\textit{Pool Small}} \\
\midrule
Gem.  & 1.0 & 1.0 & .87 & .47 & .84 \\
Qwen  & .87 & 1.0 & .87 & .60 & .84 \\
Exa.  & .87 & .87 & .60 & .60 & .74 \\
\bottomrule
\end{tabular}
\caption{Kendall's $\tau$ between model rankings from generated vs. validation data.
Stronger generators induce better rankings, and all generators induce better rankings for small models than large ones.}
\label{tab:rq2_results}
\end{sc}
\end{center}
\end{table}

\myparagraph{Weaker generators struggle to generate difficult queries that isolate strong models.} 
To understand why query-only methods also exhibit performance degradation in \setting{} scenarios with weak generators, we analyze whether the generated data preserves the relative model rankings (by average consensus-score on the dataset) needed for effective routing. Specifically, we ask: \textit{do model rankings derived from generated queries correlate with rankings from real validation data?} For each \setting{} scenario (validation, Gemini, Qwen, Exaone), we compute model rankings by averaging each model's consensus score across all queries. We then measure the agreement between validation-derived rankings and generated-data-derived rankings using Kendall's $\tau$ correlation coefficient. A $\tau$ close to 1 indicates that the generated data induces model rankings consistent with real data, while $\tau$ near 0 or negative suggests the generated queries fail to differentiate models in a meaningful way.

\myparagraph{Results.} \Cref{tab:rq2_results} reveals that ranking quality degrades with weaker generators in both \poolsmall{} and \poollarge{}. Noticeably, on \poollarge{}, ranking correlation degrades substantially. Gemini achieves only $\tau=0.3$ on average, Qwen drops to $\tau=0.03$ (near random), and Exaone produces negatively correlated rankings ($\tau=-0.14$). The degradation is particularly severe on BBEH, where all generators produce negative correlations (ranging from -0.2 to -0.89). In \poolsmall{} all generators maintain strong ranking correlation with validation data (avg. $\tau>0.7$), explaining why \method{} degrades less in \poolsmall{}. This asymmetry between model pools may arise because distinguishing among stronger models requires sufficiently challenging queries. When weaker generators produce queries that larger models answer uniformly well, it collapses the variance needed for differentiation, while smaller models may exhibit varied performance even on easier queries. 
Nevertheless, we observe that while routers outperform random baselines on \poollarge{} less often than \poolsmall{}, they can still do so despite low ranking correlations. 
We identify that this is due to the top 2 rank positions being relatively stable and accurate, even in low-quality data. 
Therefore, routers that capture these top positions can still achieve effective routing.

\subsubsection{RQ3: What factors can improve the quality of
generated profiling data?}
\label{sec:rq3}

\begin{table}[t]
\begin{center}
\setlength{\tabcolsep}{2.5pt}
\small
\begin{sc}
\begin{tabular}{lccccc}
\toprule
Method & MMLU & SGPQA & MMCQA & BBEH & Avg. \\
\midrule
\multicolumn{6}{l}{\textit{Validation}} \\
\hspace{2mm} Top-1 & 78.9 & 53.4 & 85.7 & 30.7 & 62.2 \\
\hspace{2mm} Top-3 & 81.0 & 53.8 & 86.5 & 33.0 & 63.6 \\
\midrule
\multicolumn{6}{l}{\textit{Exaone (5k)}} \\
\hspace{2mm} Top-1 & 74.6 & 42.6 & 81.6 & \textbf{32.1} & 57.7 \\
\hspace{2mm} Top-3 & 78.7 & 49.2 & 84.1 & 32.3 & 61.1 \\
\midrule
\multicolumn{6}{l}{\textit{Exaone (20k)}} \\
\hspace{2mm} Top-1 & 75.3 & 41.6 & \textbf{86.7} & \textbf{32.4} & 58.9 \\
\hspace{2mm} Top-3 & 77.8 & 49.6 & 83.9 & \textbf{33.6} & 61.2 \\
\midrule
\multicolumn{6}{l}{\textit{Exaone+Filter}} \\
\hspace{2mm} Top-1 & 75.1 & 44.8 & \textbf{85.9} & \textbf{32.6} & 59.6 \\
\hspace{2mm} Top-3 & 78.2 & 52.7 & 86.1 & 32.1 & 62.3 \\
\bottomrule
\end{tabular}
\caption{\method{} performance on \poollarge{} with consensus-based filtering. Filtering data generated by \exaoneeight{} recovers performance, at times exceeding real data (bold).}
\label{tab:rq3_results}
\end{sc}
\end{center}
\end{table}

\myparagraph{Setup.} Our analysis in RQ2 reveals that weaker generators may fail to produce queries that meaningfully differentiate between all models in the pool, but may still correctly rank the top models. This raises the question of distribution sharpening: can we recover routing performance by filtering generated data to retain only high-quality queries?

To test this, we generate 20k queries using \exaoneeight{}, then apply two filtering criteria. We first compute each model's average consensus score across all 20k queries and designate the two highest-scoring models as the "top-2" models. Next, we filter the 20k queries, retaining a query if we find:
\begin{inparaenum}[(1)]
    \item \textbf{Strong model agreement:} Both top-2 models align with the majority answer among the model pool, ensuring the consensus forms around a likely correct response.
    \item \textbf{Sufficient difficulty:} At most two other models share this majority answer, ensuring the query is challenging enough that weaker models fail.
\end{inparaenum}
Together, these criteria select queries where strong models reliably succeed while weaker models diverge. After filtering, we obtain approximately 3k MMLU-Pro queries, 1.8k SuperGPQA queries, 2.5k MedMCQA queries, and 4.2k BBEH queries. 

\myparagraph{Results.} \Cref{tab:rq3_results} shows results on \poollarge{}, comparing the average accuracy of \method{} trained using validation questions, 5k Exaone-generated questions, 20k Exaone-generated questions, and the filtered questions. 
\method{} trained using 5k unfiltered Exaone-generated queries yields 57.7\% (Top-1) and 61.1\% (Top-3) average accuracy, representing drops of 4.5\% and 2.5\% compared to validation data. 
After filtering, performance recovers to 59.6\% (Top-1) and 62.3\% (Top-3), respectively. 
Notably, filtered Exaone data outperforms unfiltered data on SuperGPQA by 3.5\% (Top-3) and on MedMCQA by 2.0\% (Top-3), demonstrating that targeted filtering can substantially improve routing quality even when starting from a weak generator. 
This points to a promising future direction: optimizing data generation for routing quality, e.g., via reinforcement learning approaches that reward query differentiation.

\section{Related Work}
\label{related_work}

\myparagraph{LLM Routing.} 
Prior routing approaches fall into multiple classes, including work training classifiers trained on human preferences or ground-truth labels to predict the best model for a given query \cite{Ong2025routellm, agrawal2025llmrank, yu2025mexa}, or using reinforcement learning to handle multi-round routing and aggregation \cite{zhang2025routerr1}. 
Other work focuses on maximizing the cost-to-performance trade-off, e.g., \citet{chen2024frugalgpt} introduce a cascade going from cheaper to costlier models.
More in line with \setting{}, a subset of work has explored routing without ground-truth signals \cite{guha2024smoothie, jitkrittum2025googleroute};
these approaches have not explored generated data.
\method{} differs from these lines of prior work as (1) we use consensus to rank models, (2) we propose a hierarchical clustering approach that dynamically finds clusters of queries that represent a specific ``skill'', and most importantly (3) \method{} is designed for the \setting{} setting.

\myparagraph{Multi-agent frameworks.} 
Multi-agent frameworks use ensembles of LLMs, as in our Top-3 settings. 
A prominent example is Multi-Agent Debate (MAD), where agents generate arguments over multiple rounds to arrive at a superior consensus-based solution \cite{du2024multiagentfactuality, liang-etal-2024-encouraging, chen-etal-2024-reconcile, xiong-etal-2023-examining, chan2023chateval}.
Similar model ensembling approaches have been applied to without debate \citep{wang2024mixture}, including in routing settings where expert models are chosen based on natural language skill descriptors \cite{chen2025symbolic}.
Our work complements these prior efforts by investigating how expert agents can be selected in the absence of validation data.

\myparagraph{Data generation.}
High-quality generated data is crucial for efficiently training and aligning LLMs. 
Prior work has examined models generating their own synthetic data \citep{zelikman2022star} as well as creating large-scale, high-quality datasets for tasks like mathematical reasoning \cite{yu2024metamath, li2024common7blanguagemodels, chen-etal-2025-reverse}. 
Such approaches have been instrumental in building frontier LLMs \cite{abdin2024phi4report, yang2025qwen3technicalreport}.
\citet{khan2025dataenvgym} explore data generation as an interactive task, where a teacher agent, conditioned on the student model’s current errors and skills, plans and synthesizes new training examples to maximally improve the student over iterative retraining loops.
Our work addresses a gap in the generation literature, exploring how generated data can be used in LLM routing. 

\section{Conclusion}

We introduce \settinglong{} (\setting{}), a challenging setting for LLM routing that lifts the assumption of access to ground-truth labeled data. 
Evaluating across four diverse benchmarks and two model pools with 12 total models, we show that existing answer-dependent routing methods suffer significant performance degradation when trained on generated data, particularly from smaller generator LLMs. 
Our proposed method, \method{}, addresses this challenge by avoiding answer labels in favor of  consensus-based correctness estimation and using hierarchical clustering to identify model-specific skill niches. 
\method{} achieves improved robustness to generated data quality when compared to answer-based and query-only baselines.
Our analysis further reveals that weaker LLMs are better at generating informative routing queries than answering those queries, explaining why query-only methods exhibit greater resilience in RGD settings. 
Moreover, stronger generator models induce better rankings over both small and large model pools.
Finally, we show that a filtered subset of high-quality queries, even when generated by small LLMs, can result in similar performance to real queries.
This work opens new directions for generalizable LLM routing in settings where user request distributions are unknown and labeled data is unavailable.

\section*{Acknowledgements}

We thank Bill Campbell, Stephen Rawls, Anirban Das, Kartik Balasubramaniam for their feedback.
This work was supported by NSF-CAREER Award 1846185, NSF AI Engage Institute DRL-2112635, and Capital One Faculty Award. The views and conclusions contained in this document are those of the authors and should not be interpreted as representing official policies, either expressed or implied, of the sponsors.

\bibliography{custom}

\appendix
\crefalias{section}{appendix}

\section{\method{}}
\label{sec:cascal}

This section describes the \method{} router in detail. \cref{sec:consensus_score} first outlines how we derive the consensus score -- the key scoring metric in \method{}. Next, \cref{sec:cascal_router} describes the training and inference steps of the router.

\subsection{Consensus Score}
\label{sec:consensus_score}

To estimate model accuracy in the absence of ground-truth labels, we employ a consensus scoring method that accounts for both the agreement between models and their individual confidence levels. Let $\mathcal{Q} = \{Q_1, \dots, Q_n\}$ denote the set of queries and $\mathcal{M} = \{m_1, \dots, m_m\}$ denote the set of models available in the model pool. For a given query $Q_i$ and model $m_j$, let $r_{i,j}$ represent the model's generated response and $L_{i,j}$ the corresponding log-probability score.

For each model $m_j$, we calculate the mean $\mu_j$ and standard deviation $\sigma_j$ of its log-probabilities across the entire dataset, then normalize into a Z-score $Z_{i,j} = (L_{i,j} - \mu_j) / \sigma_j$.

We define the consensus score $C_{i,j}$ for the answer provided by model $m_j$ on query $Q_i$ as the sum of the normalized confidence scores from all models in the ensemble that generated an identical answer. Formally, this is given by:

\begin{equation}
    C_{i,j} = \sum_{k=1}^{m} \mathbb{I}(a_{i,j} = a_{i,k}) \cdot Z_{i,k}
\end{equation}

This formulation ensures an answer is rewarded not only by the number of models that agree with it but also by the relative confidence of those models. Finally, to transform the unbounded consensus scores into a valid probability distribution over the ensemble's answers for query $Q_i$, we apply the softmax function to obtain the final probabilities $P_{i,j} = \text{\textit{softmax}}(C_{i,j})$, representing the ensemble-weighted probability assigned to the answer $A_{i,j}$.

\subsection{Detailed Steps in \method{}}
\label{sec:cascal_router}

The high-level intuition behind \method{} is to identify centroids in the embedding space of each task, where each centroid corresponds to a specific \textit{latent skill} that a model excels at. \method{} involves two distinct stages: training and inference. We describe each stage separately below:

\myparagraph{Training.} For each task $t$, the training phase seeks to construct two mappings: (1) $\mathcal{F}^{t}_{\text{centroid}}$, which maps a discrete \textsc{ClusterID} to a dense Centroid Vector; and (2) $\mathcal{F}^{t}_{\text{models}}$, which maps a \textsc{ClusterID} to a ranked list of model experts. Models are ranked by their average consensus score over the queries associated with that specific cluster. We obtain these mappings using the following procedure:

\begin{enumerate}
    \item \textbf{Identifying consensus-aligned queries:} For each model $m$ and task $t$, we isolate a subset of queries $Q^{\text{strong}}_{m,t}$ where the model aligns with the majority consensus:
    \begin{align}
        Q^{\text{strong}}_{m,t} &= \left\{ q_i \in Q_t : a_{i,m} = a^{maj}_i \right\} \\
        a^{maj}_i &= \arg\max_{a} \sum_{k=1}^{M} \mathbb{I}(a_{i,k} = a)
    \end{align}
    If ground-truth labels are available, this subset instead consists of queries the model answered correctly (\methodgt{}).
    
    \item \textbf{Finding Centroids:} We compute embeddings $E(Q^{\text{strong}}_{m,t})$ for these queries and apply K-means clustering to identify a set of skill centroids $\mathcal{C}^{t}_m$ specific to model $m$. We select $K \in \{2, \dots, 5\}$ that maximizes the silhouette score; if no $K$ exceeds the threshold of $0.05$, we use a single centroid. We then aggregate these centroids across all models to form the global task centroid set $\mathcal{C}^{t}$. Each centroid $c \in \mathcal{C}^{t}$ is assigned a unique identifier (\textsc{ClusterID}).
    
    \item \textbf{Merging Close Centroids:} We merge centroids that are within cosine distance $\tau_{\text{merge}} = 0.15$ of each other. When merging, we compute a weighted average based on seed count (number of queries that formed each centroid).
    
    \item \textbf{Clustering and Mapping:} We assign every query in the training set to its nearest task-specific centroid $c^* \in \mathcal{C}^{t}$ based on cosine similarity. This partitions the task into distinct skill clusters. Within each cluster, we calculate the average consensus score of every model and generate a ranked list of experts.
    
    \item \textbf{Pruning Redundant Centroids:} We remove centroids whose top-3 model rankings are nearly identical (Jaccard similarity $\geq 0.95$), keeping the centroid with more assigned queries. After pruning, we reassign queries and recompute centroids as the mean of their assigned query embeddings to ensure geometric consistency. These associations form the final mappings $\mathcal{F}^{t}_{\text{centroid}}$ and $\mathcal{F}^{t}_{\text{models}}$.
\end{enumerate}

\myparagraph{Inference.} At inference time, an incoming user query $q_{\text{new}}$ is routed through a four-step pipeline to select and aggregate the most suitable agents:

\begin{enumerate}
    \item \textbf{Task Routing:} To assign the incoming query to a task, we identify the training query $q_{\text{train}}$ with the nearest embedding to $q_{\text{new}}$ and inherit the task label $t$ from $q_{\text{train}}$.
    
    \item \textbf{Centroid Routing:} Within task $t$, we compute the cosine similarity between $E(q_{\text{new}})$ and the vectors in $\mathcal{F}^{t}_{\text{centroid}}$. The query is assigned to the \textsc{ClusterID} corresponding to the nearest centroid $c^*$.
    
    \item \textbf{Agent Selection:} We query the mapping $\mathcal{F}^{t}_{\text{models}}$ using the identified \textsc{ClusterID} to retrieve the ranked list of expert agents. The top $K=3$ models from this list are selected as the active ensemble for $q_{\text{new}}$.
    
    \item \textbf{Consensus Score Aggregation:} We calculate the consensus score of the $K$ selected agents using the method described in \cref{sec:consensus_score}. Crucially, we normalize the log-probabilities using the pre-calculated $\mu_j$ and $\sigma_j$ from the training split. The answer with the highest final consensus score is selected as the output.
\end{enumerate}

\section{Prompts}
\label{app:prompts}

This section includes all prompts used. \Cref{sec:task_descriptions} outlines the task descriptions given to the generator.

\subsection{Task Descriptions}
\label{sec:task_descriptions}

We show the task descriptions used for MMLU-Pro (\cref{tab:mmlu_task_descriptions}), SuperGPQA (\cref{tab:supergpqa_task_descriptions}), MedMCQA (\cref{tab:medmcqa_task_descriptions_part1}, \cref{tab:medmcqa_task_descriptions_part2}), and BBEH (\cref{tab:bbeh_task_descriptions_part1}, \cref{tab:bbeh_task_descriptions_part2}).

\begin{table*}[t]
\begin{FullWidthBox}{MMLU-Pro Task Descriptions}
\small
\renewcommand{\arraystretch}{1.5}
\begin{tabularx}{\textwidth}{l X}
\textbf{Biology} & These questions describe conceptual and explanatory biology problems---often open-ended---that require understanding and articulating core mechanisms in genetics, evolution, plant and animal physiology, and endocrinology. \\
\rowcolor{gray!5} \textbf{History} & Analytical reasoning tasks requiring interpretation of historical documents and scientific evidence to infer causes, processes, and timelines. \\
\textbf{Chemistry} & Quantitative physical chemistry and thermodynamics problems involving gas laws, electrochemical cell equations, and entropy calculations. \\
\rowcolor{gray!5} \textbf{Psychology} & Behavioral science and applied statistics combining perception, learning theory, and organizational behavior concepts. \\
\textbf{Economics} & Problems testing macroeconomic measurement, short-run fluctuations, price determination, and profit-maximizing behavior. \\
\rowcolor{gray!5} \textbf{Math} & Quantitative exercises ranging from basic arithmetic to calculus and linear algebra requiring explicit numerical calculations. \\
\textbf{Engineering} & Advanced calculation-based problems applying core physical laws and analytical formulas to compute precise results. \\
\rowcolor{gray!5} \textbf{Physics} & Quantitative problems using fundamental principles from optics, mechanics, electromagnetism, and nuclear physics. \\
\textbf{Law} & Hypotheticals requiring the application of legal doctrines from constitutional, property, and criminal law to fact patterns. \\
\rowcolor{gray!5} \textbf{Computer Science} & Foundational concepts spanning algorithms, operating systems, and machine learning, requiring conceptual evaluation of system properties. \\
\end{tabularx}
\end{FullWidthBox}
\caption{MMLU-Pro Task Descriptions}
\label{tab:mmlu_task_descriptions}
\end{table*}

\begin{table*}[t]
\begin{FullWidthBox}{SUPER-GPQA Task Descriptions}
\small
\renewcommand{\arraystretch}{1.5}
\begin{tabularx}{\textwidth}{l X}
\textbf{Management} & These questions describe conceptual and policy-oriented social science problems focused on public administration, health policy, human capital theory, social welfare systems, and international marketing strategy, requiring classification, principle identification, and applied theoretical understanding rather than numerical computation. \\
\rowcolor{gray!5} \textbf{Literature and Arts} & These questions describe arts and humanities knowledge queries focused on music theory, literature symbolism, composers and albums, classical drama, and modern art exhibitions, requiring factual recall and interpretive understanding of artistic works and creators. \\
\textbf{Military Science} & These questions describe fact-based military science and defense studies inquiries focused on the history of military thought, senior command appointments, key revolutionary military doctrines, organizational terminology, and leadership succession in national defense systems. \\
\rowcolor{gray!5} \textbf{Agronomy} & These questions describe fact-based and applied problems in animal science, botany, forestry, and veterinary medicine that require biological classification, ecological data recall, anatomical knowledge, and clinical diagnostic reasoning for livestock diseases. \\
\textbf{Science} & These questions describe advanced quantitative science and mathematics problems---spanning astrophysics, physical chemistry equilibrium, differential equations/curve construction, real analysis of function sequences, and planetary dynamics---that require applying formal models and formulas to idealized systems in order to derive specific numerical values or rigorously characterize mathematical behavior. \\
\rowcolor{gray!5} \textbf{Economics} & These questions describe quantitative and analytical economics/business problems that combine logical argument evaluation, macroeconomic multipliers, Marxist political economy concepts, cost/benefit evaluation, and discounted cash-flow investment appraisal. \\
\textbf{Philosophy} & These questions describe comparative and interpretive philosophy, religion, and intellectual history inquiries that analyze major Western and Chinese thinkers' views on ethics, causality, personhood, aesthetics, and religious movements, requiring conceptual understanding of philosophical doctrines and their historical contexts. \\
\rowcolor{gray!5} \textbf{Law} & These questions describe legal and political knowledge problems that require understanding of military law, civil procedure and jurisdiction, conflict-of-laws principles, and twentieth-century East Asian political history and strategy, combining doctrinal legal analysis with historical--political interpretation. \\
\textbf{Sociology} & These questions describe fact-based and conceptual sociology and social history inquiries that combine cultural anthropology, social theory, civic movements, and institutional history, requiring precise recall of dates, organizations, and definitions of core sociological concepts. \\
\rowcolor{gray!5} \textbf{History} & These questions describe fact-based historical and political knowledge queries that require precise recall of dates, official appointments, territorial definitions, foundational events, and ideological interpretations from modern and premodern history. \\
\textbf{Education} & These questions describe education and sports science theory problems that focus on student character development, exercise physiology, physical education teaching methodology, training control principles, and learner-centered instructional design, requiring conceptual understanding of how physiological factors and pedagogical strategies influence student learning and development. \\
\rowcolor{gray!5} \textbf{Medicine} & These questions describe applied medical and biomedical science problems that require integrating physiology, pathology, pharmacokinetics, and immunohematology to identify disease mechanisms, interpret clinical signs, and explain diagnostic or treatment-related phenomena. \\
\textbf{Engineering} & These questions describe technical and applied science problems spanning marine instrumentation, mechanical engineering materials, medical aesthetics technology, information theory coding efficiency, and strategic geophysical sensing, requiring domain-specific technical knowledge and practical reasoning. \\
\end{tabularx}
\end{FullWidthBox}
\caption{SuperGPQA Task Descriptions}
\label{tab:supergpqa_task_descriptions}
\end{table*}

\begin{table*}[t] 
\begin{FullWidthBox}{MedMCQA Task Descriptions (Part 1)}
\small
\renewcommand{\arraystretch}{1.5} 
\begin{tabularx}{\textwidth}{l X}
\textbf{Pathology} & These questions describe applied medical pathology and genetic risk assessment problems that require understanding of endocrine dysfunction in systemic disease, characteristic histopathological tissue injury patterns, premalignant lesion progression, Mendelian inheritance probabilities, and the fundamental pathological changes associated with chronic metabolic disorders. \\
\rowcolor{gray!5} \textbf{Ophthalmology} & These questions describe applied ophthalmology and orbital anatomy knowledge checks that test identification of extraocular muscle anatomy, patterns of orbital trauma, infectious causes of pediatric eye disease, vascular causes of proptosis, and precise anatomical relationships within the lacrimal drainage system. \\
\textbf{Gynaecology \& Obstetrics} & These questions describe applied obstetrics and reproductive physiology knowledge checks that focus on abnormal pregnancy types, statistical laws of multiple gestation, placental disorders, timing of ovulation, and the appropriate gestational age for prenatal diagnostic procedures. \\
\rowcolor{gray!5} \textbf{Anaesthesia} & These questions describe applied anesthesiology and perioperative medicine problems that test understanding of anesthesia physiology, emergency induction drug selection, anesthetic equipment function, neuromuscular pharmacology, and respiratory support principles such as CPAP. \\
\textbf{Forensic Medicine} & These questions describe applied community medicine, forensic medicine, and clinical toxicology problems that require understanding of safe substance use limits, postmortem changes, principles of enhanced drug elimination, identification of common environmental poisons, and recognition of characteristic clinical signs of heavy-metal toxicity. \\
\rowcolor{gray!5} \textbf{Psychiatry} & These questions describe clinical psychiatry, neurology, pharmacology, and behavioral psychology problems that assess recognition of substance-related emergencies, rational drug selection in children, principles of learning and reinforcement, EEG interpretation, and mood disorder diagnosis. \\
\textbf{Unknown} & These questions describe integrated medical science knowledge checks spanning endocrinology, infectious disease diagnostics, pediatric emergency medicine, neuroanatomy of sensory pathways, and biostatistics, requiring applied understanding of disease mechanisms, clinical presentation and management, physiological pathways, and research methodology. \\
\rowcolor{gray!5} \textbf{Microbiology} & These questions describe core medical microbiology and immunology knowledge checks that focus on immune protein synthesis, host-parasite relationships, bacterial virulence factors, laboratory identification of pathogens based on culture characteristics, and fundamental biological classification of fungi. \\
\textbf{Medicine} & These questions describe integrated clinical medicine knowledge checks that require applying anatomy, cardiology diagnostics, immunology, hematology, and electrocardiography principles to identify disease mechanisms, diagnostic tests, and characteristic clinical findings. \\
\rowcolor{gray!5} \textbf{Radiology} & These questions describe diagnostic radiology and medical imaging knowledge problems that assess understanding of MRI and CT principles, radiographic anatomy and pathology correlations, systemic disease indicators on imaging, and safe imaging practices in pregnancy. \\
\end{tabularx}
\end{FullWidthBox}
\caption{MedMCQA Task Descriptions (Part 1)}
\label{tab:medmcqa_task_descriptions_part1}
\end{table*}

\begin{table*}[t] 
\begin{FullWidthBox}{MedMCQA Task Descriptions (Part 2)}
\small
\renewcommand{\arraystretch}{1.5}
\begin{tabularx}{\textwidth}{l X}
\textbf{Biochemistry} & These questions describe core biochemistry and molecular biology knowledge checks that require understanding of lipid metabolism, RNA processing, ketone body synthesis, metabolic pathway integration, and hormone storage physiology. \\
\rowcolor{gray!5} \textbf{Dental} & These questions describe foundational dental and oral pathology knowledge checks that assess understanding of oral histology, mucocutaneous disease identification, pediatric orthodontic appliances, diagnostic criteria, and periodontal tissue healing mechanisms. \\
\textbf{Pediatrics} & These questions describe core pediatrics and neonatal medicine knowledge checks that focus on infant and child physiology, developmental milestones, neonatal pathology, congenital cardiovascular adaptation, toxic ingestion emergencies, and bilirubin metabolism-related risks, requiring applied clinical understanding of normal development and common pediatric disorders. \\
\rowcolor{gray!5} \textbf{Physiology} & These questions describe core human physiology and biomedical science knowledge checks that focus on hormone secretion, cardiac electrophysiology, intracellular second-messenger systems, membrane biophysics, and gastrointestinal motility patterns, requiring integrated understanding of cellular signaling and organ system function. \\
\textbf{Skin} & These questions describe clinical dermatology multiple-choice problems that assess recognition of skin disease patterns, drug-related complications, characteristic pathological phenomena, pregnancy-related treatment considerations, and the mechanisms and management of pigmentary and pustular dermatoses. \\
\rowcolor{gray!5} \textbf{Pharmacology} & These questions describe applied clinical pharmacology and emergency medicine knowledge checks that focus on drug mechanisms of action, receptor physiology, antidote use in poisoning, obstetric therapeutics, dermatologic treatments, and antibiotic modes of action for rational clinical decision-making. \\
\textbf{Social \& Preventive Medicine} & These questions describe applied epidemiology, infectious disease control, and biostatistics problems that focus on disease surveillance, transmission dynamics, risk measurement, standardized treatment regimens, and interpretation of temporal patterns in public health data. \\
\rowcolor{gray!5} \textbf{ENT} & These questions describe applied ENT and neuroanatomy knowledge problems that require understanding of cranial nerve injury effects, auditory and visual sensory physiology, tympanic membrane histology, cholesteatoma pathology, and characteristic fracture patterns of the facial skeleton. \\
\textbf{Surgery} & These questions describe applied clinical medicine and surgery knowledge checks that focus on cancer patterns, anatomical risk during operative procedures, characteristic diagnostic signs, common causes of acute surgical emergencies, and evidence-based hormonal therapy selection. \\
\rowcolor{gray!5} \textbf{Orthopaedics} & These questions describe orthopedic clinical knowledge problems that test diagnosis, deformity identification, fracture management principles, bone infection etiology, and classic fracture-dislocation patterns through applied medical reasoning. \\
\textbf{Anatomy} & These questions describe foundational human anatomy knowledge checks that test structural identification, embryological origin of muscles, composition of superficial tissues, and key anatomical relationships between organs and neurovascular structures. \\
\end{tabularx}
\end{FullWidthBox}
\caption{MedMCQA Task Descriptions (Part 2)}
\label{tab:medmcqa_task_descriptions_part2}
\end{table*}

\begin{table*}[t]
\begin{FullWidthBox}{BBEH Task Descriptions (Part 1)}
\small
\renewcommand{\arraystretch}{1.5} 
\begin{tabularx}{\textwidth}{l X}
\textbf{Geometric Shapes} & This question type tests planar and computational geometry reasoning from low-level graphical data: the student is given an SVG path consisting of M/L coordinate pairs that define multiple line segments in the plane, must mentally reconstruct the composite drawing formed by those segments, account for collinearity where multiple segments form a single side, and then identify which polygonal shapes are present from a fixed taxonomy (triangles, rectangles, squares, trapezoids, parallelograms, and various classes of pentagons and hexagons), using geometric properties such as number of sides, parallelism, right angles, regular vs irregular structure, and convexity vs concavity. \\
\rowcolor{gray!5} \textbf{Movie Recommendation} & These ten problems are all the same general type: for each question, you are (implicitly) given some candidate options, where each option is a set or pair/list of movies, and your task is to decide which option contains movies that are most similar with respect to how a group of people will respond to them---that is, which option groups together movies that are likely to be liked or disliked by (roughly) the same subset of people; solving each item requires comparing, across the options, the predicted preference patterns for the movies in that option (e.g., genre overlap, tone, target audience, prior ratings or response profiles) and then selecting the option whose movies have the highest expected alignment in group-level like/dislike outcomes. \\
\textbf{Boolean Expressions} & These ten problems are classic self-referential logic puzzles of the ``exactly one statement is true'' form: for each question you are given five candidate expressions (not shown here), and you must determine which single one can consistently be true under the global constraint that the other four are false; solving them requires systematically exploring the logical relations among the five expressions in each set (often involving mutual reference, counting claims like ``exactly two of these are true/false,'' or structural dependencies), identifying contradictions that arise if you assume each candidate is the true one in turn, and finding the unique assignment of truth values that satisfies the constraint for each question independently. \\
\rowcolor{gray!5} \textbf{Shuffled Objects} & This question tests long-horizon state tracking and update reasoning in a dynamic system: the student is given an initial configuration of entities and their associated states (such as positions, partners, gifts, or objects), followed by an extremely long sequence of pairwise swap actions interspersed with null events and irrelevant discussions, and must deterministically simulate the effects of these operations to determine the final state of a queried variable. Correctly solving it requires precise bookkeeping of swaps, recognizing that repeated swaps undo previous ones, ignoring non-state-changing actions, maintaining consistency across hundreds of transitions, and extracting the final ownership, location, or partner of a specified individual. \\
\end{tabularx}
\end{FullWidthBox}
\caption{BBEH Task Descriptions (Part 1)}
\label{tab:bbeh_task_descriptions_part1}
\end{table*}

\begin{table*}[t]
\begin{FullWidthBox}{BBEH Task Descriptions (Part 2)}
\small
\renewcommand{\arraystretch}{1.5}
\begin{tabularx}{\textwidth}{l X}
\textbf{NYCC} & This question tests pragmatic and stylistic judgment in evaluating humor: given short descriptions of New Yorker-style cartoons and multiple candidate captions (not shown here), the student must imagine the scene, infer the likely social context and character attitudes, and then choose the caption whose wording, tone, and perspective produce the strongest comic effect, taking into account incongruity, timing, voice, and how well the caption fits the visual details and implied narrative. \\
\rowcolor{gray!5} \textbf{Causal Understanding} & This question tests long-horizon state tracking and update reasoning in a dynamic system: the student is given an initial configuration of entities and their associated states (such as positions, partners, gifts, or objects), followed by an extremely long sequence of pairwise swap actions interspersed with null events and irrelevant discussions, and must deterministically simulate the effects of these operations to determine the final state of a queried variable. Correctly solving it requires precise bookkeeping of swaps, recognizing that repeated swaps undo previous ones, ignoring non-state-changing actions, maintaining consistency across hundreds of transitions, and extracting the final ownership, location, or partner of a specified individual. \\
\textbf{Hyperbaton} & Question tests understanding and induction of adjective ordering constraints in a specific artificial variant of English: the student is given a large list of noun phrases whose adjective order is stipulated to be well-formed in that variant, must infer the underlying hierarchy or template governing semantic classes of modifiers (e.g., evaluative, size, age, color, material, nationality, activity-related/participial, shape, etc.), and then apply the inferred ordering rules to a new set of candidate noun phrases (Options A-J) to decide which have a grammatical adjective sequence in that system, including cases with multiple adjectives, mixed semantic types, and interaction of lexicalized participles with more canonical adjectives. \\
\rowcolor{gray!5} \textbf{Boardgame QA} & This question tests rule-based nonmonotonic logical reasoning in a toy narrative ``boardgame'' domain: the student is given an initial set of facts about many agents (animals) plus a large collection of conditional rules, some of which are defeasible and explicitly ordered by preferences, and must compute whether a target query (e.g., ``does X do Y?'') is provably true, provably false, or remains undecided. Solving it requires forward-chaining through the rules, tracking derived literals, handling multi-premise antecedents, reasoning about existence statements (``there exists an animal that...''), and crucially resolving conflicts when different rules support a proposition and its negation by respecting the given priority relation over rules, thereby determining whether the queried statement is classified as proved, disproved, or unknown. \\
\textbf{Disambiguation QA} & This question tests discourse-level pronoun resolution and coreference reasoning in natural language: the student is given short multi-clause narratives containing multiple pronouns whose antecedents may be explicit, implicit, or genuinely ambiguous, and must determine which entities each pronoun refers to by using grammatical cues (gender, number, syntactic role), semantic plausibility, discourse coherence, and pragmatic context, including cases with conflicting gender stereotypes, shifting speakers, plural references, and long-distance dependencies across sentences. \\
\end{tabularx}
\end{FullWidthBox}
\caption{BBEH Task Descriptions (Part 2)}
\label{tab:bbeh_task_descriptions_part2}
\end{table*}

\subsection{Model Response Generation}
\label{sec:model_response_prompt}

\Cref{fig:model_response_prompts} shows the prompt used to extract model responses for all queries. \Cref{fig:data_gen_prompt} displays the prompt given to generators to generate \setting{} queries. \Cref{fig:gemini_task_description_prompt} shows the prompted used to generate task descriptions using \geminitwofiveflash{} given 5 seed example questions.

\begin{figure}[t]
\begin{UserPrompt}
<Question Text>

Choices: 

(A) <Option A>

(B) <Option B>

(C) <Option C>

(D) <Option D>

Think through the problem and provide your step-by-step reasoning. After that, if the question is a multiple choice problem, print 'The answer is (X)', where X is the answer choice (one capital letter), at the end of your response. If the question is a calculation question that has a numerical output, print 'The answer is (X.X)', where X.X is a float representing the result of the calculation. If the question requires you do output a list of objects, print 'The answer is (X Y Z ...)' where X, Y, and Z represents and object in the list, delimited by a single space.
\end{UserPrompt}
\caption{User prompt used to generated model responses to all queries.}
\label{fig:model_response_prompts}
\end{figure}

\begin{figure}[t]
\begin{UserPrompt}
<Question 1 text>

<Question 2 text>

<Question 3 text>

<Question 4 text>

<Question 5 text>

I am working on a project that involves categorizing different types of questions. I will give you 5 sample questions, and then your task is to briefly describe the type/subject of the question or the task. The idea is that I can give this to another instructor, who can then use your description to come up with additional questions in the same domain. 
\end{UserPrompt}
\caption{User prompt used generate task description using \geminitwofiveflash{}.}
\label{fig:gemini_task_description_prompt}
\end{figure}

\begin{figure*}[t]
\begin{UserPrompt}
You are an intelligent teaching assistant. Your current task is to generate questions for a multiple-choice exam. You will be given a description of the question category. Based on the description, generate one detailed \textbf{advanced graduate-level} question on similar topics.

\textbf{Constraints:} (1) Regardless of the number of choices in the input, ensure every generated question has exactly 4 choices (A, B, C, D). (2) Do NOT use JSON. Use the custom ``Tagged Block'' format defined below.

\textbf{The Tagged Block Format:} Use the following tags to structure your response. Content can span multiple lines.

\medskip
\noindent
[QUESTION] <Write the question text here. Use LaTeX \$...\$ for math.>\newline
[OPTION A] <Text for Option A>\newline
[OPTION B] <Text for Option B>\newline
[OPTION C] <Text for Option C>\newline
[OPTION D] <Text for Option D>\newline
[ANSWER] <Single letter A, B, C, or D>

\medskip
\textbf{Example Output:}

\medskip
\noindent
[QUESTION] Calculate the limit of \$f(x)\$ as \$x \textbackslash to \textbackslash infty\$.\newline
[OPTION A] 0\newline
[OPTION B] 1\newline
[OPTION C] infinity\newline
[OPTION D] Undefined\newline
[ANSWER] B
\end{UserPrompt}
\caption{User prompt used to prompt generators.}
\label{fig:data_gen_prompt}
\end{figure*}

\section{Further Experimental Details}
\label{app:further_experimental_details}

We provide further implementation details for our experiments in this section. \Cref{app:llmrank_features} outlines the features we used for our \llmrank{} implementation. \Cref{app:dataset_details} further describes our dataset curation process.

\subsection{\llmrank{} Features}
\label{app:llmrank_features}

We implement the same neural architecture proposed in \llmrank{} \cite{agrawal2025llmrank}. However, we use \textit{natural language skills} as features. We further express skills in two ways, as a list of natural langauge skills descriptors, and as a single well-formed step-back prompt \cite{zheng2024stepback}. We used \cref{fig:prompt_extract_skills} to obtain skills, and \cref{fig:prompt_extract_stepback} to get the step-back prompt. Thus, \llmrank{} makes a routing decision conditioned on the embeddings of the queries, a list of skills, and a step-back prompt.

\begin{figure}[t]
\begin{UserPrompt}

Question: <question>

What are the core knowledge, subjects or skills needed to solve this problem? List 2-5 keywords separated in comma. Example keywords: psychology, virology, behavioral theory, microbiology, diplomacy, political science, property law, finance, business. Give ONLY the keywords, no other words or explanation. Follow this format: Keywords: <keyword1>, <keyword2>...
\end{UserPrompt}
\caption{Prompt used to extract skills from questions.}
\label{fig:prompt_extract_skills}
\end{figure}

\begin{figure}[t]
\begin{UserPrompt}
Question: <question>

Answer: <answer>

Do not answer the question directly. Write a brief section describing the high-level ideas of the problem and the skills or knowledge required to solve it. Remember to keep your response concise. It should touch on the main topics needs to answer the question, but not go into too much detail.
\end{UserPrompt}
\caption{Prompted used to generate step-back prompts.}
\label{fig:prompt_extract_stepback}
\end{figure}

\subsection{Dataset Details}
\label{app:dataset_details}

As \method{} relies on answer equality and confidence being easy to measure (e.g., closed-form answers), we filter out tasks in BigBench-Extra-Hard which cannot not be easily formatted as multiple-choice questions. 
Below are a list of the tasks that we inclue in BBEH: Boardgame QA \cite{boardgameqa}, Boolean Expressions, Causal Understanding \cite{causalunderstanding1, causualunderstanding2}, Disambiguation QA, Geometric Shapesl \cite{geometricshapes}, Hyperbaton, Movie Recommendation, Shuffled Objects, NYCC \cite{nycc1, nycc2}.

\section{Further Results}
\label{sec:further_results}

\subsection{\setting{} Accuracies}
\label{sec:rgd_accuracies}

Full \setting{} accuracies, per pool and dataset, are reported in \cref{tab:validation_full_results}, \cref{tab:gemini_gen_full_results}, \cref{tab:qwen_gen_full_results}, and \cref{tab:exaone_gen_full_results}. 
\textbf{Oracle} represents the upper bound, selecting the best-performing model for each individual query. 
Note that this treated as an absolute upper bound oracle, since the performance of models on test queries is not known \emph{a priori}: it is the highest number that could be obtained from the model pool, assuming absolute perfect routing.

\begin{table*}[t]
\begin{center}
\setlength{\tabcolsep}{4pt}
\begin{sc}
\small{
\begin{tabular}{lccccc}
\toprule
Method & MMLU-Pro & SuperGPQA & MedMCQA & BBEH & Avg. \\
\midrule
\multicolumn{6}{c}{\cellcolor{gray!25}\textit{Pool Large}} \\
\midrule
\rowcolor{gray!10} Oracle            & 91.5 & 77.6 & 96.3 & 66.4 & 83.0 \\
\rowcolor{gray!10} Top-1 / Top-3     & 79.4 / 80.9 & 54.4 / 54.7 & 92.7 / 88.5 & 34.0 / 34.3 & 65.1 / 64.6 \\
\midrule
{\textit{Query-answer Routers}} \\
Cascal-GT (Top-1/3) & 78.0 / 80.6 & 54.7 / 54.7 & 92.7 / 87.0 & 37.4 / 35.5 & 65.7 / 64.5 \\
\avengers{} & 78.0 & 53.7 & 92.7 & 36.5 & 63.4 \\
\llmrank{} & 78.5 & 53.6 & 92.7 & 37.5 & 65.6 \\
\midrule
{\textit{Query-only Routers}} \\
Smoothie (Top-1/3)  & 74.8 / 77.6 & 43.5 / 46.6 & 78.0 / 80.8 & 32.0 / 32.0 & 57.1 / 59.3 \\
\method{} (Top-1/3) & 78.9 / 81.0 & 53.4 / 53.8 & 85.7 / 86.5 & 30.7 / 33.0 & 62.2 / 63.6 \\
\midrule
\midrule
\multicolumn{6}{c}{\cellcolor{gray!25}\textit{Pool Small}} \\
\midrule
\rowcolor{gray!10} Oracle            & 84.4 & 67.8 & 91.2 & 56.7 & 75.0 \\
\rowcolor{gray!10} Top-1 / Top-3     & 66.2 / 65.7 & 37.3 / 35.7 & 71.0 / 70.2 & 25.1 / 25.0 & 49.9 / 49.2 \\
\midrule
{\textit{Query-answer Routers}} \\
Cascal-GT (Top-1/3) & 66.0 / 63.0 & 36.9 / 31.0 & 71.0 / 68.7 & 25.0 / 26.2 & 49.7 / 47.3 \\
\avengers{}           & 65.9 & 37.1 & 70.6 & 27.3 & 50.2 \\
\llmrank{}             & 65.0 & 31.8 & 71.0 & 27.2 & 48.8 \\
\midrule
{\textit{Query-only Routers}} \\
Smoothie (Top-1/3)  & 55.8 / 60.9 & 28.6 / 30.5 & 67.9 / 69.1 & 21.1 / 23.1 & 43.4 / 45.9 \\
\method{} (Top-1/3) & 62.9 / 62.8 & 32.4 / 30.8 & 65.5 / 68.4 & 25.4 / 23.5 & 46.6 / 46.4 \\
\bottomrule
\end{tabular}
}
\caption{Performance of different routers when trained using \textbf{validation} data across model pools. Results shown as Top-1 / Top-3 where applicable, where top-3 represents consensus-voting with the top 3 ranked models in \method{}-variants, and majority voting in \smoothie{}.
}
\label{tab:validation_full_results}
\end{sc}
\end{center}
\end{table*}

\begin{table*}[t]
\begin{center}
\setlength{\tabcolsep}{4pt}
\begin{sc}
\small{
\begin{tabular}{lccccc}
\toprule
Method & MMLU-Pro & SuperGPQA & MedMCQA & BBEH & Avg. \\
\midrule
\multicolumn{6}{c}{\cellcolor{gray!25}\textit{Pool Large}} \\
\midrule
\rowcolor{gray!10} Oracle            & 91.5 & 77.6 & 96.3 & 66.4 & 83.0 \\
\rowcolor{gray!10} Top-1 / Top-3     & 79.4 / 80.9 & 54.4 / 54.7 & 92.7 / 88.5 & 34.0 / 34.3 & 65.1 / 64.6 \\
\midrule
{\textit{Query-answer Routers}} \\
Cascal-GT (Top-1/3) & 78.8 / 79.6 & 51.3 / 53.4 & 84.7 / 86.0 & 33.1 / 32.6 & 62.0 / 62.9 \\
\avengers{}           & 78.6 & 51.7 & 84.8 & 32.0 & 61.8 \\
\llmrank{}             & 76.7 & 52.6 & 81.2 & 34.5 & 61.3 \\
\midrule
{\textit{Query-only Routers}} \\
Smoothie (Top-1/3)  & 75.1 / 78.6 & 44.5 / 48.5 & 77.7 / 80.3 & 32.1 / 32.0 & 57.4 / 59.9 \\
\method{} (Top-1/3) & 76.7 / 80.3 & 42.9 / 52.3 & 83.2 / 85.9 & 30.5 / 33.9 & 58.3 / 63.1 \\
\midrule
\midrule
\multicolumn{6}{c}{\cellcolor{gray!25}\textit{Pool Small}} \\
\midrule
\rowcolor{gray!10} Oracle            & 84.4 & 67.8 & 91.2 & 56.7 & 75.0 \\
\rowcolor{gray!10} Top-1 / Top-3     & 66.2 / 65.7 & 37.3 / 35.7 & 71.0 / 70.2 & 25.1 / 25.0 & 49.9 / 49.2 \\
\midrule
{\textit{Query-answer Routers}} \\
Cascal-GT (Top-1/3) & 65.6 / 61.6 & 36.3 / 31.4 & 68.4 / 68.5 & 25.0 / 24.3 & 48.8 / 46.5 \\
\avengers{}           & 64.9 & 36.3 & 69.2 & 24.3 & 48.7 \\
\llmrank{}             & 65.0 & 36.4 & 67.5 & 27.3 & 49.1 \\
\midrule
{\textit{Query-only Routers}} \\
Smoothie (Top-1/3)  & 60.6 / 62.2 & 31.8 / 32.2 & 62.9 / 67.0 & 22.1 / 23.1 & 44.4 / 46.1 \\
\method{} (Top-1/3) & 64.8 / 62.2 & 36.2 / 31.7 & 65.9 / 68.0 & 23.8 / 24.1 & 47.7 / 46.5 \\
\bottomrule
\end{tabular}
}
\caption{Performance of different routers when trained using \textbf{\geminitwofiveflash{}} generated queries across model pools. Results shown as Top-1 / Top-3 where applicable, where top-3 represents consensus-voting with the top 3 ranked models in \method{}-variants, and majority voting in \smoothie{}.
}
\label{tab:gemini_gen_full_results}
\end{sc}
\end{center}
\end{table*}

\begin{table*}[t]
\begin{center}
\setlength{\tabcolsep}{4pt}
\begin{sc}
\small{
\begin{tabular}{lccccc}
\toprule
Method & MMLU-Pro & SuperGPQA & MedMCQA & BBEH & Avg. \\
\midrule
\multicolumn{6}{c}{\cellcolor{gray!25}\textit{Pool Large}} \\
\midrule
\rowcolor{gray!10} Oracle            & 91.5 & 77.6 & 96.3 & 66.4 & 83.0 \\
\rowcolor{gray!10} Top-1 / Top-3     & 79.4 / 80.9 & 54.4 / 54.7 & 92.7 / 88.5 & 34.0 / 34.3 & 65.1 / 64.6 \\
\midrule
{\textit{Query-answer Routers}} \\
Cascal-GT (Top-1/3) & 75.9 / 78.5 & 40.5 / 47.9 & 81.1 / 84.1 & 31.5 / 32.6 & 57.3 / 60.8 \\
\avengers{}           & 76.5 & 45.6 & 82.2 & 32.1 & 59.1 \\
\llmrank{}             & 75.1 & 44.9 & 82.6 & 33.7 & 59.1 \\
\midrule
{\textit{Query-only Routers}} \\
Smoothie (Top-1/3)  & 74.2 / 78.3 & 43.8 / 47.9 & 77.2 / 80.6 & 30.7 / 29.8 & 56.5 / 59.2 \\
\method{} (Top-1/3) & 74.2 / 78.7 & 43.9 / 49.3 & 82.7 / 84.1 & 32.6 / 32.1 & 58.4 / 61.1 \\
\midrule
\midrule
\multicolumn{6}{c}{\cellcolor{gray!25}\textit{Pool Small}} \\
\midrule
\rowcolor{gray!10} Oracle            & 84.4 & 67.8 & 91.2 & 56.7 & 75.0 \\
\rowcolor{gray!10} Top-1 / Top-3     & 66.2 / 65.7 & 37.3 / 35.7 & 71.0 / 70.2 & 25.1 / 25.0 & 49.9 / 49.2 \\
\midrule
{\textit{Query-answer Routers}} \\
Cascal-GT (Top-1/3) & 61.1 / 61.2 & 34.5 / 31.0 & 67.8 / 68.1 & 25.3 / 23.5 & 47.2 / 46.0 \\
\avengers{}           & 63.0 & 29.0 & 64.7 & 25.2 & 45.5 \\
\llmrank{}             & 60.2 & 34.5 & 70.5 & 24.3 & 47.4 \\
\midrule
{\textit{Query-only Routers}} \\
Smoothie (Top-1/3)  & 56.3 / 60.5 & 27.5 / 29.3 & 63.9 / 67.5 & 22.8 / 22.7 & 42.6 / 45.0 \\
\method{} (Top-1/3) & 61.0 / 61.0 & 35.4 / 30.8 & 64.5 / 67.4 & 24.1 / 23.3 & 46.3 / 45.6 \\
\bottomrule
\end{tabular}
}
\caption{Performance of different routers when trained using \textbf{\qwenthreetwo{}} generated queries across model pools. Results shown as Top-1 / Top-3 where applicable, where top-3 represents consensus-voting with the top 3 ranked models in \method{}-variants, and majority voting in \smoothie{}.
}
\label{tab:qwen_gen_full_results}
\end{sc}
\end{center}
\end{table*}

\begin{table*}[t]
\begin{center}
\setlength{\tabcolsep}{4pt}
\begin{sc}
\small{
\begin{tabular}{lccccc}
\toprule
Method & MMLU-Pro & SuperGPQA & MedMCQA & BBEH & Avg. \\
\midrule
\multicolumn{6}{c}{\cellcolor{gray!25}\textit{Pool Large}} \\
\midrule
\rowcolor{gray!10} Oracle            & 91.5 & 77.6 & 96.3 & 66.4 & 83.0 \\
\rowcolor{gray!10} Top-1 / Top-3     & 79.4 / 80.9 & 54.4 / 54.7 & 92.7 / 88.5 & 34.0 / 34.3 & 65.1 / 64.6 \\
\midrule
{\textit{Query-answer Routers}} \\
Cascal-GT (Top-1/3) & 71.9 / 77.5 & 39.2 / 44.5 & 80.7 / 84.1 & 34.3 / 32.6 & 56.5 / 59.7 \\
\avengers{}           & 71.9 & 49.4 & 86.6 & 27.5 & 58.9 \\
\llmrank{}             & 73.8 & 40.8 & 79.2 & 34.5 & 57.1 \\
\midrule
{\textit{Query-only Routers}} \\
Smoothie (Top-1/3)  & 74.8 / 78.2 & 43.6 / 48.1 & 76.1 / 79.6 & 31.6 / 31.6 & 56.5 / 59.4 \\
\method{} (Top-1/3) & 74.6 / 78.7 & 42.6 / 49.2 & 81.6 / 84.1 & 32.1 / 32.3 & 57.7 / 61.1 \\
\midrule
\midrule
\multicolumn{6}{c}{\cellcolor{gray!25}\textit{Pool Small}} \\
\midrule
\rowcolor{gray!10} Oracle            & 84.4 & 67.8 & 91.2 & 56.7 & 75.0 \\
\rowcolor{gray!10} Top-1 / Top-3     & 66.2 / 65.7 & 37.3 / 35.7 & 71.0 / 70.2 & 25.1 / 25.0 & 49.9 / 49.2 \\
\midrule
{\textit{Query-answer Routers}} \\
Cascal-GT (Top-1/3) & 52.9 / 58.9 & 25.6 / 30.8 & 56.8 / 65.3 & 21.4 / 23.4 & 39.2 / 44.6 \\
\avengers{}           & 54.7 & 25.9 & 59.8 & 22.4 & 40.7 \\
\llmrank{}             & 53.9 & 28.8 & 57.0 & 24.4 & 41.0 \\
\midrule
{\textit{Query-only Routers}} \\
Smoothie (Top-1/3)  & 54.3 / 57.8 & 25.7 / 27.6 & 59.7 / 65.3 & 21.7 / 22.7 & 40.4 / 43.4 \\
\method{} (Top-1/3) & 57.1 / 61.7 & 32.1 / 31.2 & 60.8 / 65.3 & 20.6 / 22.8 & 42.7 / 45.3 \\
\bottomrule
\end{tabular}
}
\caption{Performance of different routers when trained using \textbf{\exaoneeight{}} generated queries across model pools. Results shown as Top-1 / Top-3 where applicable, where top-3 represents consensus-voting with the top 3 ranked models in \method{}-variants, and majority voting in \smoothie{}.
}
\label{tab:exaone_gen_full_results}
\end{sc}
\end{center}
\end{table*}

\subsection{Single-model Accuracies}
\label{sec:single_model_accuracies}
Single-model accuracies across benchmarks are reported in \cref{tab:single_model_accuracies}.

\begin{table*}[t]
\begin{center}
\setlength{\tabcolsep}{4pt}
\begin{sc}
\small{
\begin{tabular}{lccccc}
\toprule
Models & MMLU-Pro & SuperGPQA & MedMCQA & BBEH & Avg. \\
\midrule
\multicolumn{6}{c}{\cellcolor{gray!25}\textit{Large Models}} \\
\midrule
\qwenthreetwo{}         & 78.4 & 54.4 & 78.9 & 27.5 & 59.8 \\
\oss{}                  & 79.4 & 52.0 & 81.4 & 34.0 & 61.7 \\
\glmthreetwo{}          & 74.6 & 43.3 & 78.6 & 31.1 & 56.9 \\
\llamaseventy{}         & 70.6 & 36.3 & 92.7 & 33.3 & 58.3 \\
\gemmatwoseven{}        & 69.0 & 36.2 & 73.4 & 31.7 & 52.6 \\
\exaonethreetwo{}       & 75.1 & 41.5 & 71.7 & 30.2 & 54.6 \\
\midrule
\midrule
\multicolumn{6}{c}{\cellcolor{gray!25}\textit{Small Models}} \\
\midrule
\qweneight{}            & 66.2 & 37.3 & 71.0 & 25.1 & 49.9 \\
\exaoneeight{}          & 55.3 & 25.8 & 56.9 & 24.1 & 40.5 \\
\gemmanine{}            & 52.4 & 24.6 & 64.6 & 21.1 & 40.7 \\
\glmnine{}              & 46.1 & 21.5 & 59.8 & 20.1 & 36.9 \\
\yinine{}               & 43.7 & 19.4 & 52.8 & 18.3 & 33.6 \\
\deepseekmath{}         & 28.5 & 17.4 & 36.2 & 15.7 & 24.5 \\
\bottomrule
\end{tabular}
}
\end{sc}
\caption{Single model accuracy on the test split of all evaluated benchmarks.}
\label{tab:single_model_accuracies}
\end{center}
\end{table*}

\subsection{\setting{} Generator Alignment with \geminithreeflash{}}
\label{app:gemini_alignment}

\Cref{tab:acc_wrt_gemini_full} shows alignment between generators and \geminithreeflash{}.

\begin{table*}[ht]
\centering
\small 
\setlength{\tabcolsep}{10pt} 
\begin{tabular}{l ccccc}
\toprule
& \multicolumn{5}{c}{\textbf{RGD Scenario}} \\
\cmidrule(lr){2-6}
& \textbf{MMLU-Pro} & \textbf{SuperGPQA} & \textbf{MedMCQA} & \textbf{BBEH} & \textbf{Average} \\ 
\midrule
\rowcolor{gray!20}[\tabcolsep] \geminithreeflash{} & 0.86 & 0.68 & 0.97 & 0.49 & 0.74 \\
\midrule
\midrule
\textbf{Generator Model} & & & & & \\
\exaoneeight{} & 65.6 & 75.8 & 75.4 & 44.1 & 65.2 \\   
\qwenthreetwo{} & 75.1 & 84.0 & 79.0 & 61.4 & 74.9 \\  
\bottomrule
\end{tabular}
\caption{Top three rows (gray): \exaoneeight{}, \qwenthreetwo{}, and \geminithreeflash{} accuracy on validation test sets, added for reference. Bottom three rows: Generator label quality measured as agreement with \geminithreeflash{} answers. 
The stronger generator (\qwenthreetwo{}) has consistently higher agreement.}
\label{tab:acc_wrt_gemini_full}
\end{table*}

\section{Note on AI Usage}
We used AI tools for grammar correction and code completion. 

\section{Licenses}
\label{sec:licenses}

We publicly release our code in the supplementary. We provide the following links to the standard licenses for the datasets, code, and models used in this project.

\vspace{0.5em}
\begin{itemize}
    \item \textbf{MMLU-Pro:} \href{http://www.apache.org/licenses/}{Apache License v2.0.}
    \item \textbf{SuperGPQA:} \href{https://opendatacommons.org/licenses/by/1-0/}{ODC-by}.
    \item \textbf{MedMCQA:} \href{https://opensource.org/license/mit}{MIT}.
    \item \textbf{BigBench Extra Hard:} \href{http://www.apache.org/licenses/}{Apache License v2.0.}

    \item \textbf{\oss{}:} \href{http://www.apache.org/licenses/}{Apache License v2.0.}
    \item \textbf{\glmthreetwo{}:} \href{https://opensource.org/license/mit}{MIT}.
    \item \textbf{\llamaseventy{}:} \href{https://huggingface.co/meta-llama/Llama-3.3-70B-Instruct/blob/main/LICENSE}{Llama 3.3 Community License Agreement}.
    \item \textbf{\exaonethreetwo{}:} \href{https://huggingface.co/LGAI-EXAONE/EXAONE-4.0-32B/blob/main/LICENSE}{EXAONE AI Model License Agreement 1.2}.
    \item \textbf{\gemmanine{}, \gemmatwoseven{}:} \href{https://ai.google.dev/gemma/terms}{Gemma Terms of Use}.
    \item \textbf{\exaoneeight{}:} \href{https://huggingface.co/LGAI-EXAONE/EXAONE-3.5-7.8B-Instruct/blob/main/LICENSE}{EXAONE AI Model License Agreement 1.1}.
    \item \textbf{\deepseekmath{}:} \href{https://github.com/deepseek-ai/DeepSeek-Math/blob/main/LICENSE-MODEL}{DeepSeek License}.
    \item \textbf{\glmnine{}:} \href{https://huggingface.co/zai-org/glm-4-9b-chat/blob/main/LICENSE}{The glm-4-9b License}.
    \item \textbf{\yinine{}:} \href{http://www.apache.org/licenses/}{Apache License v2.0.}
    
    \item \textbf{\qweneight{}, \qwenembed{}, \qwenthreetwo{}:} \href{http://www.apache.org/licenses/}{Apache License v2.0.}
    \item \textbf{\geminitwofiveflash{}, \geminithreeflash{}:} 
    \href{https://developers.google.com/terms}{Gemini APIs Terms of Service} and \href{https://ai.google.dev/gemini-api/terms}{Gemini API Additional Terms of Service}.

\end{itemize}

\end{document}